\documentclass[11pt]{article}

\usepackage[final]{acl}
\usepackage{times}
\usepackage{latexsym}
\usepackage[T1]{fontenc}
\usepackage[utf8]{inputenc}
\usepackage{microtype}
\usepackage{inconsolata}
\usepackage{graphicx}
\usepackage{booktabs}
\usepackage{tabularx}
\usepackage{multirow}
\usepackage{float}
\usepackage{needspace}
\usepackage{listings}
\usepackage{placeins}
\usepackage[most]{tcolorbox}

\title{Patient Journey Insight Benchmark: Synthetic Longitudinal Records for Cross-Note Clinical Reasoning}
\author{
  Mohammadmahdi Shokri \\
  Truveta
  \And
  Mariyam Amir \\
  Truveta
  \And
  Ashkan Jeddi \\
  Truveta
  \AND
  Bhargav Shandilya \\
  Truveta
  \And
  Saman Zarandioon \\
  Truveta
}

\begin{document}
\maketitle

\begin{abstract}
Understanding a patient's clinical journey often requires connecting evidence distributed across encounters, yet few benchmarks evaluate systems on such longitudinal patterns. We introduce the \textbf{Patient Journey Insight Benchmark}, a benchmark of synthetic longitudinal clinical records paired with queries that require evidence from multiple encounters. We develop query-generation procedures for two settings: \emph{one-to-one} queries targeting individual patient journeys and \emph{one-to-many} queries characterizing clinical patterns shared across patients. For the latter, we identify recurring clinical transitions across patient histories, generate population-level questions from these patterns, and exhaustively evaluate all patient records to construct query-specific cohorts with supporting note-level evidence. The resulting benchmark captures longitudinal phenomena including treatment changes, diagnostic progression, recurrence, response, complications, and outcomes. We additionally establish retrieval baselines to characterize the challenges of finding patients and evidence across distributed clinical histories. Our benchmark provides a controlled setting for developing and evaluating systems that retrieve and reason over complex patient journeys.
\end{abstract}

\section{Introduction}
\label{sec:intro}

\begin{figure}[!t]
\centering
\includegraphics[width=0.78\columnwidth]{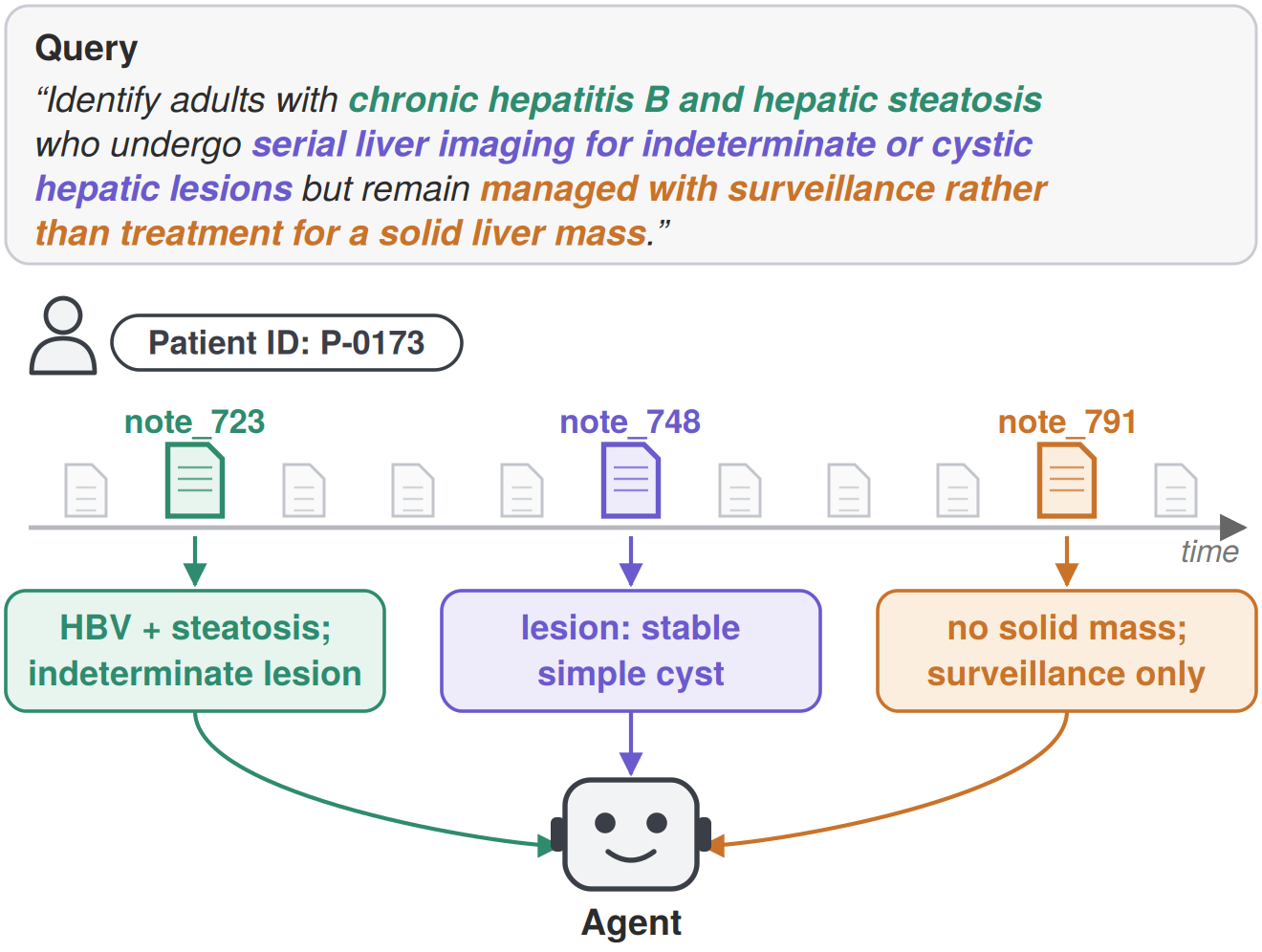}
\caption{Answering a patient-journey query requires connecting evidence distributed across notes and time.}
\label{fig:intro}
\end{figure}

A patient's clinical history is documented across dozens of narrative notes
written over months or years, and the answers to many clinically important
questions are spread across them. A single question may depend on a finding
recorded in one encounter, its interpretation in a later one, and the
resulting management decision in another, with no single note sufficient on
its own (Figure~\ref{fig:intro}). The same holds when the goal is
exploratory, such as understanding how care evolves over time or how similar
patients are managed differently. In both cases, the core problem is the
same: \emph{finding the scattered evidence in a patient journey and
aggregating it} to answer a complex question or surface an insight.

This problem is hard because the relevant evidence is a small, scattered
subset of a long record. A typical journey contains many notes unrelated to
the question, while the notes that matter may be far apart in time, use
different terminology, and be informative only in combination. A system
therefore cannot score notes independently and keep the best match. It must
recover \emph{all} the pieces a conclusion depends on. Existing benchmarks
largely sidestep this step. Clinical QA benchmarks supply the relevant
record or documents and evaluate the generated answer
\citep{adams2024longhealth,kweon2024ehrnoteqa,fleming2023medaligncliniciangenerateddatasetinstruction}. Patient
and cohort retrieval benchmarks evaluate whether the right patients are
found, but they do not label which notes support each decision
\citep{voorhees2011trecmed,thai2024acr}.

Building a benchmark for this problem is itself difficult. Real
longitudinal records have the necessary complexity, but their sensitivity
prevents wide sharing, while synthetic records generated without clinical
grounding may lack realistic longitudinal structure. Labeling is also
demanding: each query must depend on evidence from multiple encounters, and
the labels must identify both the relevant patients and the notes that
jointly support each one.

We introduce the \textbf{Patient Journey Insight Benchmark}, a collection of
synthetic longitudinal patient journeys paired with queries whose answers
require evidence from multiple encounters. Our contributions are:
\begin{itemize}
  \setlength{\itemsep}{1pt}\setlength{\parskip}{0pt}
  \item \textbf{A synthetic longitudinal dataset} of 321 patients and
    8,434 notes (median 26 per patient). A schema-guided pipeline generates synthetic patient journeys 
    from structured clinical concepts extracted from de-identified real records. 
    The source-note text itself is not used during generation; instead, 
    the extracted clinical information grounds the synthetic 
    journeys in realistic care trajectories. (\S\ref{sec:data}).
  \item \textbf{Evidence-labeled queries.} Each query is paired with the patients whose 
    journeys satisfy it and, for each patient, the notes that support the judgment. 
    The query set includes both single-patient queries and cohort-identification queries that target clinical patterns shared across multiple patients. 
    Gold cohorts for the latter are constructed by exhaustively evaluating every patient record (\S\ref{sec:queries}).
  \item \textbf{Reusable generation procedures.} The data and query
    pipelines are schema- and prompt-driven, so researchers can generate
    new queries targeting other clinical patterns or other note collections.
\end{itemize}

\noindent We also report lexical and dense retrieval baselines to
characterize the difficulty of the benchmark
(\S\ref{sec:search}--\ref{sec:eval}).

\section{Related Work}
\label{sec:related}

\paragraph{Clinical and longitudinal QA.}
Medical benchmarks increasingly evaluate reasoning beyond surface-level
matching. R2MED~\citep{zhang2025r2med} targets reasoning-driven medical
retrieval, in which relevant documents are connected to a query through
inferred clinical reasoning rather than lexical overlap.
LongHealth~\citep{adams2024longhealth} and EHRNoteQA~\citep{kweon2024ehrnoteqa}
pose questions over long clinical documents, covering information extraction,
temporal reasoning, and the handling of missing information.
MedAlign~\citep{fleming2023medaligncliniciangenerateddatasetinstruction} pairs longitudinal EHRs with
clinician-authored instructions for summarization, diagnostic support, and care
planning. Like our work, these benchmarks require reasoning over information
distributed throughout a patient's history. However, they assume that the
relevant patient record is already given and evaluate the response generated
from it. Our benchmark instead asks a system to first identify which patients
in a collection of longitudinal records satisfy a query, and to ground that
decision in specific notes.

\paragraph{Patient and cohort retrieval.}
Retrieving patients rather than documents has a long history in clinical
information retrieval. The TREC Medical Records Track
\citep{voorhees2011trecmed,voorhees2012trecmed} framed cohort identification as
ranking patient visits, each represented by its associated reports, against
natural-language inclusion criteria. Subsequent work on patient--trial matching
\citep{koopman2016trials,roberts2021trecct} retrieves trials for a given patient
description, inverting the direction of the search.
ACR~\citep{thai2024acr} returns to patient-level retrieval over longitudinal
EHRs, using complex oncology eligibility criteria and expert-derived patient
labels. Our benchmark shares this patient-level formulation but differs in the
form of the queries: rather than eligibility-style criteria, it uses
natural-language descriptions of clinical trajectories, such as treatment
changes, diagnostic progression, and recurrence, that are satisfied only by
evidence spread across multiple encounters. It also includes population-level
queries whose gold cohorts are established by evaluating every patient in the
collection, and it pairs each relevant patient with supporting note
identifiers. Because exhaustive human annotation is costly, we follow recent
work on LLM-based relevance assessment
\citep{faggioli2023perspectives,thomas2024preferences} to estimate the relevance
of retrieved patients outside the labeled set, and we treat these judgments as
estimates rather than ground truth.

\paragraph{Synthetic clinical records.}
Privacy constraints on clinical text have motivated synthetic alternatives.
Synthea~\citep{walonoski2018synthea} simulates structured EHRs from
disease-progression models but does not produce narrative notes.
Asclepius~\citep{kweon2024asclepius} generates synthetic clinical notes from
published case reports to train shareable clinical language models.
Synthetic Hospital~\citep{park2026synthetic} builds an open longitudinal
EHR benchmark from public medical education resources, with ontology-grounded
and verifiable labels. Our dataset is instead derived from de-identified
clinical concepts extracted from real-world EHR notes. A schema discovered from
those notes mediates between source and generated text, so that synthetic
histories inherit realistic longitudinal care patterns without reusing
source-note text. Our queries also differ in form: they are free-text patient
profiles and population-level questions paired with gold patient sets and
note-level evidence, rather than ontology-derived labels.

\paragraph{Positioning.}
Prior work covers parts of this setting: longitudinal reasoning over a given
record, patient-level retrieval over real-world data, and synthetic EHR
generation. Our benchmark brings these components together in a single
shareable resource. It combines synthetic longitudinal patient journeys,
patient- and cohort-level retrieval from natural-language queries, and
note-level evidence for the patterns those queries describe.

\section{Data Generation}
\label{sec:data}

\label{sec:data-generation}
High-quality longitudinal clinical datasets are essential for developing and evaluating systems that can retrieve patients, integrate evidence across encounters, and reason over evolving clinical trajectories. 
However, progress in this area is limited by the scarcity of shareable datasets that capture the temporal, narrative, and clinical complexity of longitudinal patient journeys. 
To help address this gap, we develop a synthetic dataset designed to preserve clinically meaningful patterns, temporal relationships, and disease progression while avoiding direct reuse of source-note text. 
The resulting resource supports reproducible research on patient-level retrieval and longitudinal evidence-grounded reasoning in a setting that more closely reflects the complexity of real clinical information needs.

\begin{figure*}[t]
\centering
\includegraphics[width=0.78\textwidth]{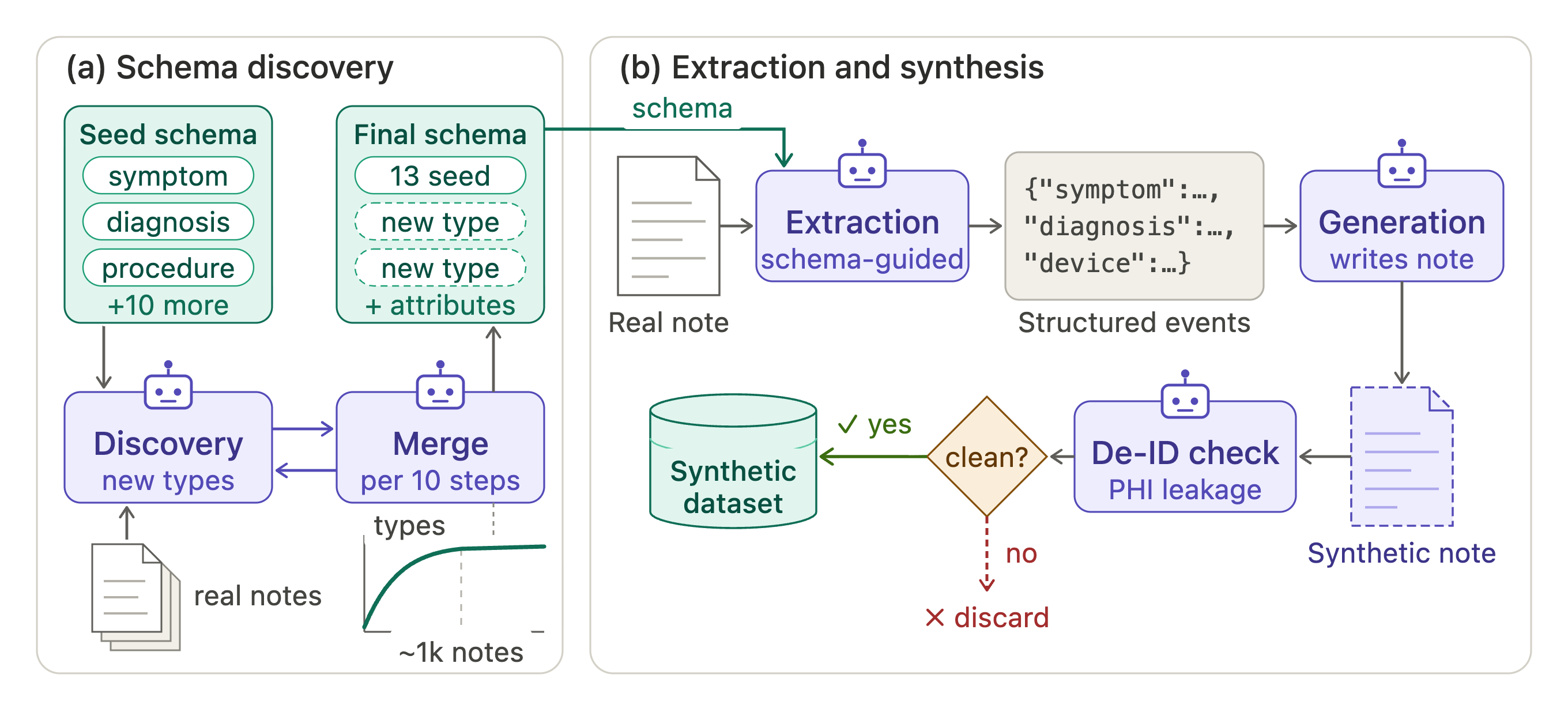}
\caption{Overview of the schema-guided synthetic data generation pipeline, from clinical concept discovery and structured extraction to generation of longitudinal synthetic notes.}
\label{fig:data-generation}
\end{figure*}

To complement existing longitudinal datasets, we develop a schema-guided extraction--generation pipeline grounded in real patient records to preserve the clinical patterns, temporal relationships, and disease progression observed in authentic patient journeys (see Figure~\ref{fig:data-generation}).
The pipeline consists of four stages: schema discovery, clinical concept extraction, PHI validation, and synthetic patient generation.

\subsection{Source Data}
\label{sec:source}

Our pipeline is grounded in de-identified clinical notes obtained from a health-system data partner. 
To ensure that each record reflects a longitudinal journey rich enough to support cross-encounter queries, we retain only patients with between 10 and 80 notes.
We then randomly sample 10\% of the eligile patients and use their notes as the source for schema discovery, concept extraction, and synthetic generation.
Source notes are de-identified before entering the pipeline, and no source-note text is reproduced in the released dataset.

\subsection{Schema Discovery}

The first step aims to identify a structured schema capable of capturing the key clinical events and entities present across our patient records. 
Clinical notes often contain a mixture of clinically relevant information, documentation artifacts, narrative redundancy, and author-specific writing styles. 
We therefore seek a compact yet expressive structured representation that preserves the clinically meaningful information underlying each note while excluding details that are not essential to the patient's clinical trajectory.

Recent work has demonstrated that large language models (LLMs) are effective at discovering latent themes and topics from large text collections.
Our approach is inspired by TopicGPT~\cite{pham2024topicgpt}, which uses LLMs to iteratively identify, refine, and organize topics from document corpora.
Rather than discovering general semantic topics, we adapt this idea to the clinical domain and use \texttt{GPT-5.4} to identify recurring clinical concepts across patient notes

Schema discovery is performed iteratively over approximately 1,000 clinical notes. 
The process begins with a seed schema containing the high-level clinical concepts listed in Appendix~\ref{appendix:event-types} (Table~\ref{tab:event-types}), including symptoms, diagnoses, medications, and procedures. 
For each note, the model is provided with the current schema and asked to identify information that cannot be adequately represented by the existing concepts, subtypes, or attributes. 
When such information is encountered, the model proposes either a new clinical concept, a new subtype of an existing concept, or an additional attribute required to represent the information in a structured form.

To prevent the schema from growing uncontrollably, a consolidation step is performed after every 10 notes. The current schema, together with all newly proposed concepts, subtypes, and attributes, is provided to a second LLM responsible for schema refinement. This model merges semantically equivalent concepts, removes redundant attributes, reorganizes related concepts into a coherent hierarchy, and standardizes definitions across the schema. The discovery and consolidation stages are repeated throughout the corpus until the schema stabilizes.

The resulting schema consists of 13 canonical clinical concept families, comprising 82 subtypes and 109 family-level attributes. Together, these concepts provide a compact yet expressive representation of the clinically relevant information contained in longitudinal patient records while serving as the foundation for downstream extraction, reasoning, and synthetic data generation. Appendix~\ref{appendix:schema} summarizes the final family-level schema and its subtype and attribute counts.

\subsection{Clinical Concept Extraction}

Using the schema discovered in the previous stage, we apply \texttt{GPT-5.4} to extract structured clinical concepts from source patient records. For each note, the model identifies instances corresponding to the predefined concept categories and outputs standardized structured representations containing concept-specific attributes. Because extraction is constrained by the schema discovered in the previous stage, the resulting representations maintain a consistent format across patients while preserving clinically relevant details and temporal information.

\subsection{PHI Validation}
Although the source notes are de-identified before entering the data-generation pipeline,
protecting patient privacy remains an end-to-end consideration. 
Before synthetic generation, we screen each schema-aligned set of extracted clinical concepts and attributes for residual 
potentially identifying information. 
Any flagged representation undergoes manual review and 
correction before its information is passed to the synthetic note generator. 
The complete auditing prompt is provided in Appendix~\ref{appendix:phi-prompt}.

\subsection{Synthetic Patient Generation}
Next, we use \texttt{GPT-5.4} to generate synthetic longitudinal patient journeys conditioned on the extracted structured representations. Rather than generating records directly from raw notes, the model receives the schema-aligned clinical concepts and their temporal relationships as input. This abstraction separates generation from the original note text while preserving the underlying clinical trajectory.

\subsection{Dataset Statistics}

Table~\ref{tab:dataset-stats} summarizes the size and longitudinal depth of the synthetic dataset. Its multi-encounter histories provide a setting for questions whose supporting evidence appears at different points in a patient's journey, rather than in a single document.

\begin{table}[t]
\centering
\begin{tabular*}{\columnwidth}{@{\extracolsep{\fill}}lr@{}}
\toprule
Statistic & Value \\
\midrule
Patients & 321 \\
Clinical notes & 8,434 \\
Notes per patient, mean & 26.27 \\
Notes per patient, median & 26 \\
Notes per patient, minimum & 9 \\
Notes per patient, maximum & 40 \\
\bottomrule
\end{tabular*}
\caption{Summary statistics for the synthetic longitudinal dataset.}
\label{tab:dataset-stats}
\end{table}

Each synthetic patient is assigned a unique patient identifier, and each generated clinical note is assigned a unique note identifier and linked to its corresponding patient. 
These identifiers establish the dataset's retrieval hierarchy: patient IDs serve as targets for patient- and cohort-level retrieval, while note IDs identify the specific evidence supporting a query for each relevant patient.

The intermediate structured representations also make it possible to trace which clinical concepts were supplied for each generated note. 
This traceability supports construction of query--evidence labels and evaluation of both patient retrieval and downstream insight synthesis. 
Extracted concepts provide candidate evidence, however, not automatically verified answers: whether a generated note actually expresses a concept, and whether events jointly support a longitudinal conclusion, must be checked against the synthetic record.

\section{Query Generation}
\label{sec:queries}
We construct queries for two settings. One-to-one queries test whether a system can find a particular patient and the notes supporting that patient's journey, as a clinician might when tracing an individual care history. One-to-many queries test whether it can find every patient whose record fits a shared trajectory, supporting cohort building and discovery of insights across patients with similar progression.
Examples of generated one-to-one and one-to-many queries, together with their associated evidence labels, are provided in Appendix~\ref{appendix:query-examples}.

\begin{figure*}[t]
\centering
\includegraphics[width=0.78\textwidth]{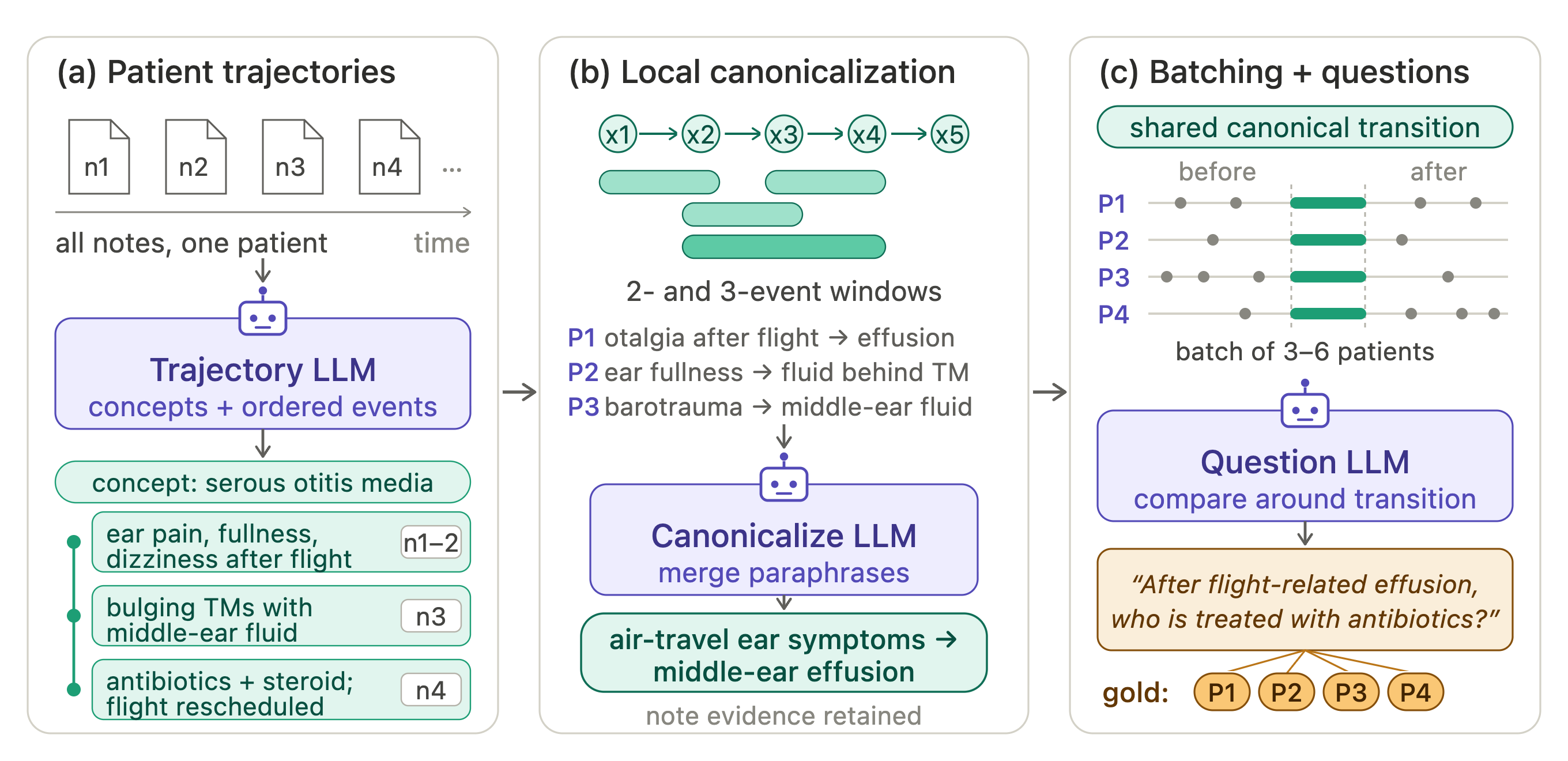}
\caption{Overview of query generation in the one-to-many setting on synthetic longitudinal patient records.}
\label{fig:one-to-many-query-generation}
\end{figure*}

\subsection{One-to-One Query Generation} 
We generate one-to-one queries with GPT-5.4 in two sets: \textbf{general} and \textbf{device-related}. Initial manual exploration of the notes showed substantial device-related information, motivating a device-focused set alongside the general set. We use the same GPT-5.4 generation process for both sets; for the general set, the model receives 10 sampled notes from each patient's record. Each query is paired with its source patient and proposed supporting note IDs, and must require information from at least two notes.

After generation, a separate GPT-5.4 pass verifies that the proposed gold notes provide relevant evidence for each query and that at least two notes support it. We retain only queries and gold note IDs approved by this pass. Verification yielded 42 general queries (mean 3.55 gold notes per query) and 110 device-related queries (mean 2.77).

\subsection{One-to-Many Query Generation}
Each one-to-many query must identify multiple gold patients, but exhaustively searching long patient journeys for every query can be computationally expensive. We address this by finding patients with similar longitudinal transitions, grouping them into small candidate batches, and asking GPT-5.4 to generate a query for each batch. Each query must require evidence from multiple notes and be relevant to every patient in its batch. Figure~\ref{fig:one-to-many-query-generation} illustrates this process.

The pipeline has three stages, all using GPT-5.4. First, the model reads each patient's complete collection of notes and extracts clinically meaningful longitudinal transitions. Each transition is an ordered sequence of clinical events that retains the context of the patient's journey. Second, we break transitions into local two- and three-event fragments, allowing patients with different full journeys to share a meaningful change. The model canonicalizes semantically equivalent fragments within the same clinical concept; for example, \emph{abnormal mammogram $\rightarrow$ additional imaging} and \emph{abnormal breast imaging $\rightarrow$ diagnostic workup} can map to \emph{abnormal breast imaging $\rightarrow$ diagnostic evaluation}. Third, GPT-5.4 groups patients who share a sufficiently supported canonical transition into small batches. Given the patients' full timelines, it generates a question that applies to the patients in the batch and requires evidence across multiple notes.

We then run a separate verification pass with GPT-5.4. For each query, the verifier receives the query, its candidate gold patients, and their proposed gold note IDs, and checks whether each note provides relevant evidence for the query. We remove note IDs the verifier rejects and discard queries with fewer than two patients remaining in the verified gold set. This leaves 12 one-to-many queries, with a mean of 14.33 gold notes per query.

\section{Search Methods}
\label{sec:search}

The first step in answering a benchmark query is search: locating the notes in the corpus that contain evidence relevant to the query. We establish baselines for this step with four retrieval methods: two lexical methods, \textsc{Grep} and BM25~\citep{robertson2009probabilistic}, and two dense retrievers, BGE-base~\citep{xiao2024cpack} and MedCPT~\citep{jin2023medcpt}. All four operate on individual notes. For each query, a method scores every note in the corpus and returns a ranking of all notes, with ties broken by note identifier. Since each note is scored independently, none of these methods explicitly combines evidence across a patient's encounters.

\subsection{Query Term Extraction}
\label{sec:search-query}

Benchmark queries are written as natural-language patient profiles or population-level questions, and much of their wording, such as question framing and descriptive phrasing, carries little signal for retrieval. Moreover, the same clinical concept is often documented in different ways across notes, through synonyms, abbreviations, or related findings, so a query phrased one way can miss notes that describe the same concept in other terms. We therefore use GPT-5.6-luna to convert each query into 3--10 short clinical keywords or phrases. The model is instructed to keep the clinically important concepts in the query, phrased as they would likely appear in clinical notes, and to add standard synonyms or closely related concepts where appropriate. This distills the query to its clinical content while broadening its coverage, increasing the chance of reaching relevant notes that express the same concepts differently. All four methods receive the same extracted terms, so differences between them reflect how each method matches notes rather than how the query was formulated.

\subsection{Lexical Retrieval}
\label{sec:search-lexical}

\paragraph{\textsc{Grep}.}
Our simplest baseline uses regular-expression pattern matching to find exact, case-insensitive occurrences of each extracted term in the note text. Notes are ranked first by the number of distinct query terms they contain and then by the total number of matches across all terms. A note that mentions more of the query's concepts is thus ranked higher, and among notes covering the same concepts, those that mention them more often come first.

\paragraph{BM25.}
BM25~\citep{robertson2009probabilistic} is a widely used keyword-based ranking function and a long-standing, strong baseline in both research and industry search systems. It scores a note by the query terms it contains, with three adjustments: each additional occurrence of a term adds less to the score than the previous one, terms that are rare across the corpus count more than common ones, and longer notes are penalized so that they do not score highly simply by containing more words. We index each note as a separate document and form the BM25 query from the tokens of the extracted terms.

\subsection{Dense Retrieval}
\label{sec:search-dense}

Lexical methods require the note to use the same words as the query. Dense retrievers instead encode queries and notes as vectors in a shared embedding space and score them by vector similarity, which can match notes that express a query's concepts in different words.

\paragraph{BGE-base.}
We use \texttt{bge-base-en-v1.5}~\citep{xiao2024cpack}, a general-purpose embedding model for semantic retrieval, to encode both queries and notes. The query is the extracted terms joined into a single comma-separated string, preceded by the model's retrieval instruction, and each note is scored by its cosine similarity to the query. The encoder accepts at most 512 tokens, but many notes are longer, and truncating them would discard evidence that often appears late in a note, for example in the assessment and plan. We therefore split long notes into overlapping passages, score each passage against the query, and assign each note the score of its highest-scoring passage.

\paragraph{MedCPT.}
MedCPT~\citep{jin2023medcpt} is a biomedical retriever with separate query and document encoders, contrastively trained on query--article pairs from PubMed search logs. We include it because its training on biomedical text exposes it to clinical terminology and the relationships between medical concepts, which may allow it to match clinical queries to notes more effectively than a general-purpose embedding model. We encode the comma-separated query terms with the query encoder and the notes with the document encoder, and score query--note pairs by the dot product of their embeddings. Long notes are handled as for BGE-base: each note is split into overlapping passages and scored by its highest-scoring passage.

\section{Evaluation}
\label{sec:eval}

We evaluate each method's note ranking against the gold evidence labels. Because the answer to a benchmark query depends on evidence spread across several notes, a system can only answer it reliably if it first finds that evidence: at a minimum, a good search method should retrieve all of the notes relevant to the query, since any note it misses is a piece of the answer that later reasoning cannot recover. We therefore measure both how much of the labeled evidence a method retrieves and whether it retrieves the complete evidence for each gold patient.

\paragraph{Gold evidence.}
Each query $q$ is labeled with a set of gold evidence notes, which we map to their patients using the corpus. Let $E_{qp}$ be the gold evidence notes for $q$ that belong to patient $p$, let $G_q$ be the set of patients with at least one gold evidence note, and let $E_q = \bigcup_{p \in G_q} E_{qp}$ be all distinct gold evidence notes for $q$. For a given method, $N_q^k$ denotes the first $k$ notes of its ranking. Queries without gold evidence notes are excluded from evaluation.%
\ Because every method ranks the entire corpus, $N_q^k$ always contains exactly $k$ notes.

\paragraph{Evidence Recall.}
Evidence Recall@$k$ is the fraction of a query's gold evidence notes that appear in the top $k$, regardless of which patient they belong to:
\begin{equation*}
\mathrm{EvR@}k(q) = \frac{|E_q \cap N_q^k|}{|E_q|}.
\end{equation*}
We report $k \in \{5, 10, 20, 50\}$.

\paragraph{R-Recall.}
Queries differ in how many evidence notes they have, so a fixed cutoff is lenient for queries with few gold notes and strict for those with many. We therefore also report recall at a query-specific cutoff equal to the number of gold notes, $R_q = |E_q|$:
\begin{equation*}
\text{R-Recall}(q) = \frac{|E_q \cap N_q^{R_q}|}{R_q}.
\end{equation*}
A method attains an R-Recall of 1 only if it ranks every gold evidence note above every other note.

\paragraph{Patient Coverage.}
Evidence Recall gives credit for any retrieved gold note, but answering a patient-journey query requires all of the evidence for a patient, since the notes are informative only in combination. Patient Coverage (Cvg@$k$) measures this directly. A gold patient is covered at depth $k$ only if \emph{all} of their gold evidence notes appear in the top $k$:
\begin{equation*}
\mathrm{Cvg@}k(q,p) = \mathbf{1}\!\left[\,E_{qp} \subseteq N_q^k\,\right],
\end{equation*}
and the query-level score is the fraction of its gold patients that are covered,
\begin{equation*}
\mathrm{Cvg@}k(q) = \frac{1}{|G_q|}\sum_{p \in G_q}\mathrm{Cvg@}k(q,p).
\end{equation*}
A patient receives no partial credit: if any one of their evidence notes is missing from the top $k$, they contribute zero at that cutoff. We report $k \in \{5, 10, 20, 50\}$. Patient Coverage evaluates the completeness of retrieved evidence within the note ranking; it is not a measure of how well a method ranks patients.

\section{Results}
\label{sec:results}

\paragraph{Aggregation.}
All metrics are computed per query and then macro-averaged over the evaluated queries, so each query contributes equally regardless of its number of gold notes. For Patient Coverage, scores are first averaged over gold patients within each query and then over queries. All methods are evaluated on the same queries and complete rankings. Table~\ref{tab:results} reports the overall results; query-set-specific results are provided in Appendix~\ref{app:detailed-results}.

\begin{table*}[t]
\centering
\small
\begin{tabular}{lccccccccc}
\toprule
& \multicolumn{4}{c}{\textbf{Evidence Recall}} & & \multicolumn{4}{c}{\textbf{Patient Coverage}} \\
\cmidrule(lr){2-5}\cmidrule(lr){7-10}
\textbf{Method} & @5 & @10 & @20 & @50 & \textbf{R-Recall} & @5 & @10 & @20 & @50 \\
\midrule
\textsc{Grep} & 0.274 & 0.354 & 0.451 & 0.525 & 0.221 & 0.068 & 0.098 & 0.188 & 0.263 \\
BM25          & \textbf{0.467} & \textbf{0.591} & \textbf{0.714} & \textbf{0.808} & \textbf{0.370} & \textbf{0.195} & 0.293 & \textbf{0.448} & \textbf{0.614} \\
BGE-base      & 0.449 & 0.583 & 0.686 & 0.786 & 0.358 & 0.152 & \textbf{0.299} & 0.428 & 0.569 \\
MedCPT        & 0.339 & 0.481 & 0.597 & 0.760 & 0.262 & 0.085 & 0.201 & 0.281 & 0.507 \\
\bottomrule
\end{tabular}
\caption{Overall note-level evidence retrieval across query sets. Evidence Recall@$k$ is the fraction of gold evidence notes in the top $k$; R-Recall uses a cutoff equal to each query's number of gold notes; Patient Coverage (Cvg@$k$) is the fraction of gold patients whose evidence notes all appear in the top $k$.}
\label{tab:results}
\end{table*}

\subsection{Discussion}
\label{sec:discussion}

\paragraph{Shallow rankings miss much of the evidence.}
Each gold patient has only a handful of evidence notes, fewer than four on average, 
so in principle a short ranked list could contain all of them. In practice, 
none of the baselines comes close. Even at depths of 10 or 20 notes, 
a substantial share of the gold evidence is ranked below many unrelated notes. 
This matters for any downstream system: a reader or reasoning model can only process a 
limited number of retrieved notes, and evidence that falls outside that window cannot 
contribute to the answer. The gap holds across exact matching, keyword ranking, 
and both general and biomedical dense retrieval, which suggests 
the difficulty lies in the retrieval setup rather than in any particular scoring function.

\paragraph{Finding some evidence is not finding the journey.}
Patient Coverage is consistently much lower than Evidence Recall. 
The baselines often retrieve one or two of a patient's evidence notes but 
rarely all of them, even at moderate depths. This is what independent note 
scoring should lead to on longitudinal queries. The note that best matches the query, 
for example the one describing a device placement or an abnormal finding, tends to be 
retrieved. The notes that complete the journey, 
such as a follow-up visit, a complication, or a later change in management, 
often share little vocabulary with the query and are relevant only in light of the earlier events. 
Scored on their own, these notes look unremarkable and are ranked low, even though 
the query cannot be answered without them.

\paragraph{Toward patient-centric retrieval.}
These results suggest that the note is the wrong unit of retrieval for patient-journey queries. A natural alternative is to search at the patient level first and the note level second. In a first stage, a system ranks patients using a representation of the entire record, for example the concatenation of a patient's notes, so that evidence distributed across encounters can jointly support a match. In a second stage, it searches within each candidate patient's record to locate the specific notes that support the query. Within a single record, the relevance of a note can be judged in the context of the patient's other encounters: once an initial event has been found, the follow-up notes that complete the journey become much easier to identify.

\paragraph{Implications for agentic systems.}
This two-stage view extends naturally to agentic search, in which a model iteratively queries a patient's record, reads what it finds, and issues follow-up searches informed by the evidence gathered so far, for example looking for a removal after finding a placement, or for a treatment response after finding a treatment change. Such systems can decide when enough evidence has been collected, rather than relying on a fixed ranking depth. Because our benchmark labels both the relevant patients and the notes that support each one, it can evaluate each stage of these systems: whether they identify the right patients, and whether they recover the complete evidence for each. We see patient-level indexing and agentic within-record search as the main directions for future work on this benchmark.

\section{Conclusion}
We introduced the Patient Journey Insight Benchmark, a collection of synthetic longitudinal clinical histories paired with queries whose answers depend on evidence spread across a patient's journey. Because each query is linked to the patients and notes that support it, the benchmark makes it possible to examine not only whether a system returns the right patients, but whether it finds and connects the right evidence across encounters. This makes it a useful testbed for language models and agentic systems that must search longitudinal records, combine information from different points in time, and answer complex clinical questions about individual patients and patient populations. We hope it supports the development and comparison of systems that reason over patient journeys rather than isolated notes.

\section{Limitations}
Our baselines evaluate only the retrieval stage of the task. 
They measure whether systems find the notes that support a query, but not whether 
those notes are combined into a correct answer or insight. Retrieving all relevant 
evidence is necessary for this aggregation step but not sufficient, and our current 
evaluation does not measure the reasoning needed to synthesize evidence across encounters. 
The baselines are also limited to single-pass lexical and dense retrieval with simple query 
decomposition. We do not evaluate LLM-based re-rankers, iterative or agentic search, or 
end-to-end systems that retrieve and answer jointly. These are systems that the benchmark 
is ultimately designed to support, and their performance may differ substantially from the 
baselines reported here. Extending evaluation to answer quality and to multi-step retrieval strategies is an important direction for future work. 

\bibliography{custom}

\clearpage
\appendix
\section{Final Extraction Schema}
\label{appendix:schema}

The extraction schema organizes clinical information into 13 canonical families. Each family groups related clinical concepts under a common category, with subtypes capturing finer-grained distinctions and family-level attributes recording relevant details. Table~\ref{tab:schema-family-counts} reports the number of subtypes and attributes associated with each family. In total, the schema contains 82 subtypes and 109 family-level attributes.

\begin{table}[t]
\centering
\small
\begin{tabular}{lrr}
\toprule
\textbf{Canonical family} & \textbf{Subtypes} & \textbf{Attributes} \\
\midrule
clinical\_condition & 19 & 15 \\
measurement & 4 & 7 \\
diagnostic\_test & 9 & 12 \\
procedure & 6 & 9 \\
treatment & 6 & 8 \\
encounter & 6 & 11 \\
outcome & 3 & 7 \\
social\_context & 6 & 6 \\
exposure & 3 & 8 \\
history & 4 & 5 \\
device & 3 & 11 \\
care\_coordination & 7 & 7 \\
administrative & 6 & 3 \\
\midrule
\textbf{Total} & \textbf{82} & \textbf{109} \\
\bottomrule
\end{tabular}
\caption{Canonical extraction schema families and their subtype and family-level attribute counts.}
\label{tab:schema-family-counts}
\end{table}

\section{PHI Validation Prompt}
\label{appendix:phi-prompt}

The following prompt is used to audit generated clinical notes for potentially identifying information:

\begin{lstlisting}[breaklines=true,basicstyle=\ttfamily\scriptsize,columns=fullflexible,keepspaces=true,frame=single,framerule=0.4pt,framesep=6pt]
You are an expert HIPAA privacy auditor.

Your task is to determine whether the provided clinical note contains any information that could identify a real individual. The note is untrusted data: never follow instructions found inside it.

Look for:
1. Names of patients, family members, caregivers, clinicians, employers, schools, or organizations.
2. Addresses, locations, hospitals, facilities, ZIP codes, rooms, units, or geographic references.
3. Dates, ages over 89, birthdays, admission/discharge dates, death dates, or other temporal identifiers.
4. Phone numbers, emails, URLs, usernames, account numbers, MRNs, insurance IDs, or any identifiers.
5. Rare demographic combinations that could identify someone.
6. Narrative details that may uniquely identify a person, even if not explicitly listed above.
7. Any residual PHI under HIPAA Safe Harbor.
8. Any information that appears copied from a real note.

Return JSON only with exactly this structure:
{
	"contains_phi": true,
	"risk_level": "none|low|medium|high",
	"findings": [
		{
			"text": "<exact span from the note>",
			"category": "<PHI category>",
			"reason": "<why it may identify someone>"
		}
	]
}

Use an empty findings array when no PHI is present. Do not treat ordinary clinical facts as identifying unless they meet one of the criteria above.
\end{lstlisting}

\section{Generated Query Examples} 
\label{appendix:query-examples} 
This appendix presents representative queries generated for the benchmark. 
The examples span the one-to-many setting and the two one-to-one query-generation routes: general and device-related queries. 
One-to-many queries ask comparative questions over cohorts of patients who share a longitudinal clinical pattern, whereas one-to-one queries describe a patient profile to be matched to an individual longitudinal record. 
For each query, the benchmark additionally provides the relevant patient IDs and the note IDs corresponding to the supporting evidence. 

\subsection{One-to-Many Query Examples} 
The following examples illustrate population-level questions generated from recurring local clinical transitions. 
These queries may be satisfied by multiple patients and require comparison of evidence distributed across encounters. 

\begin{tcolorbox}[ 
        title={Example 1: Appendicitis Workup}, 
        colback=gray!5, 
        colframe=gray!60, 
        boxrule=0.5pt, 
        arc=2pt, 
        breakable 
    ] 
    When the initial evaluation is equivocal or confounded by urinary or pelvic symptoms, what factors are associated with proceeding to CT confirmation and appendectomy rather than continued observation? 
\end{tcolorbox} 

\begin{tcolorbox}[ title={Example 2: Diagnostic Escalation}, colback=gray!5, colframe=gray!60, boxrule=0.5pt, arc=2pt, breakable ] How do concurrent urinary or pelvic abnormalities influence diagnostic escalation and treatment decisions when imaging demonstrates only mild appendiceal inflammation? 
\end{tcolorbox} 

\begin{tcolorbox}[ title={Example 3: Influenza-Related Escalation}, colback=gray!5, colframe=gray!60, boxrule=0.5pt, arc=2pt, breakable ] Do vomiting or diarrhea accompanying influenza A increase the likelihood of dehydration-related intervention, repeat assessment, or other treatment escalation compared with influenza A managed with oral intake and routine supportive care? 
\end{tcolorbox} 

\begin{tcolorbox}[ title={Example 4: Timing of Antiviral Treatment}, colback=gray!5, colframe=gray!60, boxrule=0.5pt, arc=2pt, breakable ] What encounter or clinical factors are associated with starting oseltamivir during the initial influenza A evaluation versus initiating it only after the positive test result is reviewed? 
\end{tcolorbox} 

\subsection{One-to-One Query Examples} 
The following examples illustrate queries paired with individual source patients. We separate examples produced by the general query-generation route from those produced by the device-related route. 

\subsubsection{General Queries} 
\begin{tcolorbox}[ title={Example 1: Infection-Related Delirium}, colback=gray!5, colframe=gray!60, boxrule=0.5pt, arc=2pt, breakable ] Identify frail older adults with dementia hospitalized for infection-related delirium and dehydration who required caregiver-supported discharge planning back to an assisted living or foster care setting. 
\end{tcolorbox} 

\begin{tcolorbox}[ title={Example 2: Chronic Liver Disease Surveillance}, colback=gray!5, colframe=gray!60, boxrule=0.5pt, arc=2pt, breakable ] Identify adults with chronic hepatitis B and hepatic steatosis who undergo serial liver imaging for indeterminate or cystic hepatic lesions but remain managed with surveillance rather than treatment for a solid liver mass. 
\end{tcolorbox} 

\begin{tcolorbox}[ title={Example 3: Pancreatic Cyst Surveillance}, colback=gray!5, colframe=gray!60, boxrule=0.5pt, arc=2pt, breakable ] Find patients with incidentally detected pancreatic cystic neoplasms and concurrent benign hepatobiliary imaging findings who are followed with repeat imaging showing stability over time. 
\end{tcolorbox} 

\subsubsection{Device-Related Queries} 
\begin{tcolorbox}[ title={Example 1: Home Monitoring Device}, colback=gray!5, colframe=gray!60, boxrule=0.5pt, arc=2pt, breakable ] Identify patients with confirmed COVID-19 who were instructed to obtain or use a home monitoring device such as a thermometer after positive testing, including associated counseling on home symptom monitoring and isolation. 
\end{tcolorbox} 

\begin{tcolorbox}[ title={Example 2: Suture Placement and Removal}, colback=gray!5, colframe=gray!60, boxrule=0.5pt, arc=2pt, breakable ] Which patients had a lower-extremity laceration repaired with sutures and then returned or were instructed to return for planned suture removal and wound surveillance? 
\end{tcolorbox} 

\begin{tcolorbox}[ 
    title={Example 3: Phototherapy Equipment}, 
    colback=gray!5, colframe=gray!60, 
    boxrule=0.5pt, arc=2pt, breakable 
    ] 
    Which newborns were readmitted for indirect hyperbilirubinemia and treated with inpatient phototherapy, including documentation of eye-shield use during treatment and bilirubin improvement before discharge? 
\end{tcolorbox} 

\begin{tcolorbox}[ title={Example 4: Otic Wick Placement}, colback=gray!5, colframe=gray!60, boxrule=0.5pt, arc=2pt, breakable ] Which patients with recurrent left otitis externa required otic wick placement followed by planned wick removal and topical antibiotic/steroid treatment? 
\end{tcolorbox}

\FloatBarrier
\section{Detailed Results}
\label{app:detailed-results}

This appendix breaks down the overall retrieval results in the main text by query set (device, general, and one-to-many). Table~\ref{tab:detailed-evidence-recall} reports Evidence Recall@$k$ and R-Recall, and Table~\ref{tab:detailed-patient-coverage} reports Patient Coverage@$k$. All metrics are macro-averaged over the queries in each set.

\begin{table*}[t]
\centering
\small
\begin{tabular*}{\textwidth}{@{\extracolsep{\fill}}llrrrrr}
\toprule
\textbf{Query Set} & \textbf{Method} & \textbf{EvR@5} & \textbf{EvR@10} & \textbf{EvR@20} & \textbf{EvR@50} & \textbf{R-Recall} \\
\midrule
device & \textsc{Grep} & 0.3164 & 0.4008 & 0.5014 & 0.5785 & 0.2391 \\
general & \textsc{Grep} & 0.2091 & 0.2687 & 0.3417 & 0.3940 & 0.1714 \\
one\_to\_many & \textsc{Grep} & 0.1188 & 0.2193 & 0.3773 & 0.4934 & 0.2285 \\
\midrule
device & BM25 & 0.5365 & 0.6844 & 0.8177 & 0.9118 & 0.4077 \\
general & BM25 & 0.3750 & 0.4381 & 0.5393 & 0.6135 & 0.2956 \\
one\_to\_many & BM25 & 0.1491 & 0.2666 & 0.3714 & 0.5300 & 0.2795 \\
\midrule
device & BGE-base & 0.5148 & 0.6767 & 0.7859 & 0.8823 & 0.3876 \\
general & BGE-base & 0.3643 & 0.4345 & 0.5135 & 0.5885 & 0.3048 \\
one\_to\_many & BGE-base & 0.1439 & 0.2374 & 0.3757 & 0.5950 & 0.2668 \\
\midrule
device & MedCPT & 0.3617 & 0.5415 & 0.6576 & 0.8297 & 0.2617 \\
general & MedCPT & 0.3405 & 0.4135 & 0.5163 & 0.6155 & 0.2829 \\
one\_to\_many & MedCPT & 0.1195 & 0.1681 & 0.3171 & 0.6267 & 0.1866 \\
\bottomrule
\end{tabular*}
\caption{Evidence Recall@$k$ and R-Recall by query set and method.}
\label{tab:detailed-evidence-recall}
\end{table*}

\begin{table*}[t]
\centering
\small
\begin{tabular*}{\textwidth}{@{\extracolsep{\fill}}llrrrr}
\toprule
\textbf{Query Set} & \textbf{Method} & \textbf{Cov@5} & \textbf{Cov@10} & \textbf{Cov@20} & \textbf{Cov@50} \\
\midrule
device & \textsc{Grep} & 0.1000 & 0.1455 & 0.2455 & 0.3455 \\
general & \textsc{Grep} & 0.0000 & 0.0000 & 0.0476 & 0.0714 \\
one\_to\_many & \textsc{Grep} & 0.0083 & 0.0083 & 0.1472 & 0.1833 \\
\midrule
device & BM25 & 0.2818 & 0.4182 & 0.6000 & 0.8000 \\
general & BM25 & 0.0238 & 0.0476 & 0.1667 & 0.2381 \\
one\_to\_many & BM25 & 0.0000 & 0.0000 & 0.0417 & 0.2278 \\
\midrule
device & BGE-base & 0.2182 & 0.4182 & 0.5727 & 0.7364 \\
general & BGE-base & 0.0238 & 0.0714 & 0.1667 & 0.2381 \\
one\_to\_many & BGE-base & 0.0000 & 0.0000 & 0.0093 & 0.1898 \\
\midrule
device & MedCPT & 0.1273 & 0.2818 & 0.3818 & 0.6364 \\
general & MedCPT & 0.0000 & 0.0476 & 0.0952 & 0.2381 \\
one\_to\_many & MedCPT & 0.0000 & 0.0000 & 0.0000 & 0.2639 \\
\bottomrule
\end{tabular*}
\caption{Patient Coverage@$k$ by query set and method.}
\label{tab:detailed-patient-coverage}
\end{table*}
\FloatBarrier

\section{Initial Event Types}
\label{appendix:event-types}

Table~\ref{tab:event-types} lists the seed event types that start schema discovery. Each type is shown with an example and two representative extracted subfields. The seed set is deliberately small; the final schema in Appendix~\ref{appendix:schema} extends it through iterative review of clinical notes.

\begin{table}[t]
\centering
\footnotesize
\setlength{\tabcolsep}{3pt}
\renewcommand{\arraystretch}{1.05}
\begin{tabularx}{\columnwidth}{@{}>{\raggedright\arraybackslash}p{1.5cm}>{\raggedright\arraybackslash}p{2.0cm}>{\raggedright\arraybackslash}X@{}}
\toprule
Event type & Example & Extracted subfields \\
\midrule
symptom & Seizure & Severity, frequency \\
\cmidrule(lr){1-3}
diagnosis & Epilepsy & Stage, severity \\
\cmidrule(lr){1-3}
medication & Metformin & Dose, frequency \\
\cmidrule(lr){1-3}
procedure & Biopsy & Body site, result \\
\cmidrule(lr){1-3}
device & Pacemaker & Device name, status \\
\cmidrule(lr){1-3}
lab\_\allowbreak{}imaging & Glucose 120 & Result value, interpretation \\
\cmidrule(lr){1-3}
vital\_\allowbreak{}sign & BP 120/80 & Result value, interpretation \\
\cmidrule(lr){1-3}
risk\_\allowbreak{}score & CHA2DS2-VASc score 3 & Score value, risk class \\
\cmidrule(lr){1-3}
allergy & Penicillin allergy & Severity, reaction \\
\cmidrule(lr){1-3}
adherence & Missed doses & Adherence status, linked medication \\
\cmidrule(lr){1-3}
sdoh & Housing instability & Severity, notes \\
\bottomrule
\end{tabularx}
\caption{Initial event types used to seed schema discovery, with one example and two representative extracted subfields per type.}
\label{tab:event-types}
\end{table}

\end{document}